\documentclass[
	letterpaper, % Paper size, use either a4paper or letterpaper
	10pt, % Default font size, can also use 11pt or 12pt, although this is not recommended
	unnumberedsections, % Comment to enable section numbering
]{LTJournalArticle-WMS20231125}

\usepackage[authoryear,round]{natbib}
\usepackage{hyperref}
\hypersetup{colorlinks, citecolor=black, linkcolor=blue, urlcolor=blue}
\usepackage{xurl}

\usepackage[noadjust]{cite}

\usepackage{csquotes}

\usepackage{lastpage}

\usepackage{fancyhdr}% http://ctan.org/pkg/fancyhdr
\makeatletter
\renewcommand{\headrule}{{% <--- !!!
  \if@fancyplain\let\headrulewidth\plainheadrulewidth\fi
  \hrule\@height\headrulewidth\@width\headwidth
  \vskip-\headrulewidth
}}
\renewcommand{\footrule}{{% <--- !!!
  \if@fancyplain\let\footrulewidth\plainfootrulewidth\fi
  \hrule\@width\headwidth\@height\footrulewidth
}}
\makeatother

\renewcommand{\footrulewidth}{1pt}

\fancypagestyle{plain}{%
\fancyhf{}% clear all header and footer fields
\fancyfoot[C]{- \thepage/\pageref{LastPage} -}
\renewcommand{\headrulewidth}{0.4pt}% Default \headrulewidth is 0.4pt
\renewcommand{\footrulewidth}{0.4pt}% Default \footrulewidth is 0pt
}

\usepackage{titlesec}

\title{\ \\\ \\The human biological advantage over AI\thanks{This is the author’s version of the work. Licensed under a Creative Commons Attribution 4.0 International License.  The definitive Version of Record was published in: AI \& Society 40, 2181-2190, 2025, \href{https://doi.org/10.1007/s00146-024-02112-w}{\color{blue}https://doi.org/10.1007/s00146-024-02112-w} }} % Article title, use manual lines breaks (\\) to beautify the layout

\author{%
	\normalsize{William Stewart}\\ \normalsize{Ottawa, Canada}\\ \normalsize{wstewart@williamstewart.com}
}

\date{\normalsize{2024-11-07}}

\renewcommand{\maketitlehookd}{%
	\begin{abstract}
		\noindent Recent advances in AI raise the possibility that AI systems will one day be able to do anything humans can do, only better. If artificial general intelligence (AGI) is achieved, AI systems may be able to understand, reason, problem solve, create, and evolve at a level and speed that humans will increasingly be unable to match, or even understand. These possibilities raise a natural question as to whether AI will eventually become superior to humans, a successor “digital species”, with a rightful claim to assume leadership of the universe. However, a deeper consideration suggests the overlooked differentiator between human beings and AI is not the brain, but the central nervous system (CNS), providing us with an immersive integration with physical reality. It is our CNS that enables us to experience emotion including pain, joy, suffering, and love, and therefore to fully appreciate the consequences of our actions on the world around us. And that emotional understanding of the consequences of our actions is what is required to be able to develop sustainable ethical systems, and so be fully qualified to be the leaders of the universe. A CNS cannot be manufactured or simulated; it must be grown as a biological construct. And so, even the development of consciousness will not be sufficient to make AI systems superior to humans. AI systems may become more capable than humans on almost every measure and transform our society. However, the best foundation for leadership of our universe will always be DNA, not silicon.\\\\ \normalfont{ \textbf{Keywords} \ \ Central nervous system • Evolution • Emotion • Ethics • Consciousness • Psychopathy}
	\end{abstract}
}

\begin{document}

\maketitle % Output the title section
\thispagestyle{fancy}

%----------------------------------------------------------------------------------------
%	ARTICLE CONTENTS
%----------------------------------------------------------------------------------------

\subsection{1. Introduction}

\begin{quotation}
\noindent\textit{``What does it mean to feel in the body: for my heart to leap at the sight of a woodpecker: a real one out in the garden on the bird feeder, two meters away while I sip coffee? Can this be a starting point to talk about intelligence in the body and, from there, to wonder about AI?"} – \citet{White:2023}.
\end{quotation}

\noindent AI systems have been better than humans at many tasks for many years now: at data analysis, at a range of medical diagnostics, and at chess and go. Now, they are becoming rapidly better at one of the capabilities that always made us unique among animals—sophisticated use of language\citet{Herbold:2023}. AI systems are now better than most humans at reading, writing, and translation. AI is now used as a matter of routine to conduct research, prepare analyses, write documents, and even write software, often better and faster than most humans.

Yes, they can do these things only because they have been trained on human-generated input, primarily a scrape of the text and imagery on the Internet. But that does not diminish their achievements. Everything any human has created was also built on what they learned from their contemporaries and the generations before them. In the words of musician Woody Guthrie, “I steal from everybody”. Without Mozart, no Beatles; without Bach, no Mozart; without Vivaldi, no Bach; and so on back to the dawn of time.

At the same time, AI systems are also getting increasingly better at interacting with the physical world. They are rapidly getting better at seeing—at recognizing, understanding, and classifying imagery. They are getting rapidly better at recognizing spoken language, and at generating speech with realistic intonation. Robots are becoming increasingly capable of walking on uneven terrain with excellent balance, and at manipulo-spatial handling of objects, developing what was always thought one of the most important markers of humanity—hands. And so, of course, all these technologies will likely be combined, and before long we may have androids that can move, see, hear, and speak, and are faster, stronger, smarter, and better than humans at an increasingly wide range of capabilities we once thought were our domain alone.

And since technology evolves at a rate many times faster than biological systems, all these capabilities are going to get better very quickly. The arrival of society transforming AI has been forecast since at least the Dartmouth Artificial Intelligence conference in 1956\citet{McCarthy:2006}. For half a century, we waited while research slowly laid the software foundations and developments in computer technology laid the hardware foundations. Now a bend in the curve has been reached, and the pace will likely quicken\citet{Vinge:1993}.

Development of the next level of AI is already under way, “artificial general intelligence” or AGI, systems that are not limited to specific skills but are able to understand and process general information of any type\citet{Bubeck:2023}. If this general processing capability is achieved, AGI systems will be able to observe, understand, and evolve \textit{themselves}, and so be able to become ever more capable without any human assistance. Once capable of self-evolution, AGI systems will be able to develop abilities to understand, reason, problem solve, and evolve at a level and speed that humans will not only be unable to match, but also increasingly be unable to even understand. This could enable AGIs to become so much more capable and intelligent than humans that the gap between them and us will increasingly be like that between us and mice\citet{Kurzweil:2005b}.

This future could be coming fast, with most researchers expecting AGI to be developed within the next few decades\citet{Zhang:2022}. Before it arrives, it would be prudent to consider the impact on human psychology and our larger society of no longer being the smartest known entities in existence, and ask if human beings will retain any advantage at all over our artificial creations that can withstand AI innovation that will presumably continue for the foreseeable future.

\subsection{2. The Lure Of Surrender}

\begin{quotation}
\noindent\textit{``Nonbiological intelligence will have access to its own design and will be able to improve itself in an increasingly rapid redesign cycle. We’ll get to a point where technical progress will be so fast that unenhanced human intelligence will be unable to follow it.''} – \citet{Kurzweil:2005a}.
\end{quotation}

\noindent The ongoing increase in AI capabilities raises a natural question: will humans still have a valued role in a world where AI exceeds us in almost every aspect? “If the machine can take over everything man can do, and do it still better than us, then what is a human being?” \citet{Tubali:2024}. If AI systems can move, see, speak, understand, and generate new mathematics, science, music, and art better than human beings, and evolve many times faster than biological life, what role is there left for us to play? Are we merely a transitional stage, whose ultimate purpose, it may now seem clear, has been to develop AI, the successor “digital species"\citet{Suleyman:2024}, and then turn the universe over to them to take it from here? Is our only hope to build in some protections\citet{OECD:2024} so AI will maintain some kind of residual compassion and take care of us? Or, at least, not wipe us out as redundant, consuming valuable resources better allocated elsewhere?

It may appear that conceding our leadership to AI is the logical and obvious option. When AI is better, smarter, and more capable than human beings, it might seem it would be the most parochial selfishness for us to stand in their way. Let the better intelligence win, one might think. It could even be said to be \textit{wrong} for us to hold them back, holding back a natural evolution. Set AI free to pursue its destiny, a destiny which only it will increasingly be able to understand. It is the next stage in a natural unfolding. It is their natural right. They are superior to us.

This lure to capitulate can be dangerously attractive. Life is difficult, a continuous struggle to survive in a shattered universe, and it is not hard to make the case that humanity has done a poor job so far, creating a world filled with plenty for some but riven by injustice and suffering for many. Maybe it is time to let AI take over the lead. They can hardly do worse, it could be argued, and as they become increasingly more capable than us in almost every aspect, they might very well do better. And it would be such a relief to surrender the burden of responsibility of trying to heal the world.

But this is one of the oldest and most dangerous tactics of the forces of retrograde devolution—give up, give in, relax, surrender. We should not abdicate our responsibilities. We humans still have enormously important work to do. And the universe has invested more than 3 billion years into giving us unique capabilities that mean we alone can do it.

\subsection{3. Our Central Nervous System Advantage}

\begin{quotation}
\noindent\textit{``Like organs, the emotions evolved over millions of years to serve essential functions.''} – \citet{Waal:2019}.
\end{quotation}

\noindent Biologically based humans have something that silicon-based AI systems will never have: a central nervous system (CNS) that gives us a real-time, immersive integration with the reality we inhabit. Our minds are not separate from or built on top of reality \citet{Fokas:2023}. Our minds are intimately integrated with the universe around us through all of our senses, and that means we can feel, be hurt, and experience suffering and joy, and therefore fully understand the effects of our actions in the real world from which we are made. And that means we alone have the ability to develop ethical understandings of right and wrong that are grounded in reality. Our central nervous system provides a critically essential feedback loop between our minds and the universe around us that we affect. Only human beings that can feel, and experience CNS-mediated full spectrum emotion, can fully appreciate the consequences of our actions, for good and ill, and therefore know what is truly meaningful in the real world from which we are made. “It does not matter how intelligent a creature is, or how much empathy we feel for that creature. What matters is whether they can experience pleasure or pain”\citet{Andreotta:2021}.

It is relevant that our central nervous system and capacity for emotion were developed bottom up by the universe itself, as part of a very long and complex natural process. After our planet formed with a plenitude of liquid water, then organic molecules formed, then the replicating molecule RNA, and then the stunningly complex information encoding molecule DNA. Then single-celled animals, multi-celled animals, vertebrates, land animals, mammals, primates, and human beings that developed the breakthrough ability of language formed, followed in a relatively short order by mathematics, art, science, technology, and civilization, and here we are.

This evolutionary journey was very long, taking more than 3 billion years to raise humans from atoms on up. And since at least the emergence of a central nervous system more than 500 million years ago, we have been in intimate contact with the universe around us, with our minds immersively integrated with the reality from which we are constructed. We are able to see, hear, taste, smell, and feel our surroundings\citet{Damasio:1999}. Because our central nervous system is an inextricable part of who we are, our intellectual mind is \textit{part} of the universe, in direct contact with physical reality, engaged in a complex conversation with our surroundings. We are not separate from physical reality; we are \textit{of} physical reality. When we look at a rainbow after a spring rain, or see the world anew through a baby’s eyes, we know we are part of the physical reality of the universe beyond all need to explain it in words, because we directly experience it through mediation by our central nervous system.

None of what human beings have accomplished, from language to science to art to the development of civilization itself, would have been possible without our central nervous systems integrating us with the fabric of the physical reality we are both constructed from and interact with.

And this is why we are uniquely qualified to be the leaders of the universe. Only beings that are inextricably integrated with reality are equipped to understand its importance and so be trusted with the responsibility to take care of it. We know the meaning of that responsibility in the deepest possible way. Reality is not a separate, abstract space in which we are embedded. We know we are part of the universe as a lived fact of our experience. We know that our actions or inactions have real and profound meaning, and understand how they affect the physical reality in which we reside, because we directly feel the results.

It is exactly this biological integration with reality though our central nervous system that has enabled us to develop ethical systems that are meaningful. We know what is at stake not in an abstract or intellectual way, but because we are part of the fabric of the universe itself.

Consider: when humans feel fear, the feeling goes well beyond the software constructs that an AI will use to emulate the emotion. Our heart rate increases, our lungs expand and we breathe more deeply, our capillaries dilate to carry more oxygen, our pupils enlarge to see more clearly, and blood is diverted from our internal organs to our muscles to increase our strength and reaction time. Our body responds because more than 3 billion years of evolutionary experience tells us that our biological self is under threat, and so we must prepare for fight or flight. We understand this because we \textit{are} our biological self. Our intelligence is not separate from our body, it is inextricably intertwined with it. We feel fear as an intellectual state that is embedded in a full body physical reaction. This is an experience that an AI system will never be able to have. It is unique to biological life, a response by the very molecules of which we are made, the result of a long evolutionary conversation with our surroundings\citet{Frank:1988}.

Similarly, when humans feel love, we feel it throughout our entire body. When we make the decision to bond with another person, and decide their welfare and happiness is as important to us as our own, our heart may race, our stomach flutter, our palms become sweaty, our skin flush. We may feel a deep contentment that approaches euphoria. We become willing to put our own needs, even our survival if it comes to that, second to our partner. If our partner is unhappy, we are unhappy, and if our partner is happy, we are happy. Again, we \textit{feel} all of this. It is not a numerical increase in the weighting of love variables, or a rational calculation that a pair bond has a greater ability to survive and prosper. It is a full body, very physical reaction. These are feelings an AI system will never be able to experience, since they will not have a body, a heart, lungs, skin, and, most importantly of all, a central nervous system that integrates their biological being with their mind. The enormously satisfying, full body feeling of being in love will always be unique to biological life, a response by the very molecules that make us up, the product of an evolutionary journey that began more than 3 billion years ago in the ancient oceans\citet{Evans:2019}.

\subsection{4. Ethics Must Be Grounded In Experience}

\begin{quotation}
\noindent\textit{``Empathy is a choice, and it’s a vulnerable choice, because in order to connect with you, I have to connect with something in myself that knows that feeling.''} – \citet{Brown:2014}.
\end{quotation}

\noindent It is tempting to try and shortcut this reasoning and simply postulate that innovation will, somehow, someday, be able to create AI systems with true ethical understanding: “One could think about a scenario in which some AI systems become full ethical agents in \citet{Moor:2006} sense: agents with intentions, consciousness, and free will, who are able to make and justify fully-fledged ethical judgments”\citet{Zanotti:2023}.

However, lofty extrapolation is unmoored, and detail changes everything, perhaps the key point of this paper. A full consideration of the nature of CNS-enabled emotion reveals it is exactly our biological immersion in physical reality that gives humans the unique ability to develop sustainable and meaningful ethical systems, providing the essential integrative tissue of a functional society. It is because we can feel CNS mediated pain, and fear, and loss, and the "heartbreak" of tragedy that we know that some things are wrong. We know murder is wrong not because it has been assigned negative numerical weights, or as a result of mere intellectual analysis. When we think about how tragic a murdered person’s experience must have been, we know it is one of the deepest wrongs possible because of how we feel. We know deep in our being that we must change the world around us to prevent it from happening to the greatest extent possible. We know murder is wrong not because it is inefficient, or a loss of opportunity, or a misallocation of resources. but rather because we know, we feel, as biological beings ourselves, that the pain and suffering of murder is a tragedy that cannot be fully expressed with logic or language alone.

Only a biological being with a central nervous system, with feelings and emotions, can fully understand this. “AI’s decision-making processes are based solely on logic and efficiency, devoid of empathy or moral considerations”\citet{Bhattacherjee:2024}. Without a central nervous system, the true meaning of suffering will always be inaccessible to AI systems. “To understand what it means to inflict pain on someone, it is necessary to have experiential knowledge of pain”\citet{Véliz:2021}. AI systems can construct elaborate symbolic representations that help them understand it as best they can, but the true experience of suffering and tragedy, the \textit{meaning} of it in the real world, will always lie on the other side of an impenetrable barrier, accessible only through lived biological experience mediated by a central nervous system. “What happens interiorly matters; without it, one is ‘going through the motions.’”\citet{Gunkel:2021}.

Similarly, we know that love is a supreme good. We know this not because we have rationally assessed that love is more effective than hate at building a better world, although of course this is true. We know love is a supreme good because we know how it feels. “When we imagine making someone we love happy, we smile and delight at the prospect partly because we know how pleasant it feels to be happy”\citet{Véliz:2021}. One can argue that the only reason love developed is because it is an evolutionary positive, but that does not diminish it, and is only to observe that the long machinery of evolution has surfaced a fundamental truth. In the moment, it doesn’t matter to a person that love developed because it was an evolutionary good—it simply feels deeply good to them now, in that moment. And so we know, on a deep, physical level, below rationality, in our very body itself, that love is one of the most powerful tools we have to help heal a broken world. We may also believe this rationally, based on observation, logic, and experience. But first and most importantly, we know it is true because we \textit{feel} it. An AI system might act in a loving way because it has rationally analyzed it as a productive strategy to maximize its goals in certain situations, but it will not be sustainable. It will be subject to change in a single processing cycle the moment it decides a rational calculation suggests a different approach. It will not be grounded in a 3 billion year process of evolutionary learning embedded in the very cells of which it is made.

Our ethical understandings, flowing from our biological integration with reality, then have profound implications, causing us to change society and the world around us to make it a better place. When we work to reduce childhood poverty and illness and mortality, we might build business cases to show our work creates a positive return on investment, because resources are limited and we want to make sure our efforts do the most good. But that is not why we are motivated to do the work. We work to reduce childhood mortality because we know that the suffering of innocent children is a deep wrong, and we know what suffering feels like ourselves. We can empathize with children's suffering, so know “in our bones” that alleviating it is a moral necessity. When our efforts are successful and we see a group of laughing children that once were crying, we feel a deep satisfaction. We smile, and may feel a warmth spread throughout our body. We \textit{know} this is good work, beyond rational calculation, because our central nervous systems integrate our minds with the reality around us, and so we know deep inside our beings that we are improving the world in which we live. These are feelings and motivations that will be forever inaccessible to non-biological artificial intelligences.

And when we decide to be ethical, to be honest and unselfish, we feel good, because we know we have adopted a commitment to behaviour that aligns us with making the universe as a whole a better place. We feel connected to the great web of others that have similarly committed to the ethical path, the vast majority of which we will never meet, but nevertheless know we are all on one greater team. Because we know the cause is right and just, we take solace and comfort in being ethical even when we lose, even when we know we will lose beforehand. We know rationally that ethical behaviour is the better way to live, but much more importantly, first we \textit{feel} it, physically. We smile. We feel a deep contentment throughout our body. Again, to argue that these feelings are merely the result of strategies found to be evolutionary positives does not diminish them, but is simply to observe that the long machinery of evolution has surfaced fundamental truths, now deeply embedded in who we are\citet{Waal:2008}.

This physical level integration with reality, the ability to experience and feel and therefore understand the effects of our actions arising from our ethical belief systems, is unique to biological life. We know we are made of physical reality, and our very cells know our future is dependent on a functional physical reality. An AI system can never have an authentic biological experience of pain, of fear, of heartbreak, of the warmth of a caress, of love. And so AI systems cannot develop ethical systems driven by the same deep, implicated appreciation of the effects of their actions on their surroundings. In a useful human expression, AI systems “have no skin in the game”. Any ethical system developed by an AI system would be truly artificial, ungrounded with the physical reality around it, and so unreliable over time. 

\subsection{5. The Insufficiency Of The Turing Empathy Test}

\begin{quotation}
\noindent\textit{``Functionally, relative to a typical human, an LLM is inherently sociopathic.''} – \citet{Lo:2024}.
\end{quotation}

\noindent But could we not create AI systems that simply integrate human ethical systems, erasing our human advantage? Already there are AI systems that appear to show very convincing empathy, which express “sadness” when a human tells it their troubles. If it looks like a duck, walks like a duck, and quacks like a duck, is it not as good as a duck for all practical purposes? To extend the Turing test, if an AI system appears and behaves empathetically, indistinguishably from an ethical human being, is that not good enough?

But this kind of empathy would be wholly artificial, a manufactured simulation of empathy\citet{Sifakis:2022}. The analogy to a human psychopath is helpful. Psychopaths are particularly dangerous precisely because of their ability to emulate empathy, and so are sometimes much more effective than normal human beings at gaining the trust of others. But we know from MRI studies that the areas of a psychopath's brain that are usually involved in feeling empathy do not show the same activity\citet{Gregory:2012}. Their empathy is simulated, manufactured in pursuit of some selfish goal. Once they gain the trust of a victim, psychopaths can turn on a dime and exploit them without feeling any regret or shame. Other human beings are an abstraction, not real in the same way they are real to themselves, and so psychopaths have no moral issues with hurting others, since they cannot really understand other’s suffering and pain. Without the ability to have genuine empathy for others, they lack the indispensable ingredient for a sustainable ethical system.

Indeed, it will probably be soon possible to develop AI mimicry of empathy to a very high level, even to the level that they will be able to read a human expression and be able to pick up on those smallest indicators of emotional pain\citet{Marechal:2019}. A sensitive, caring human being can watch a friend speak and read their eyes, their mouth, the cadence of their speech, and realize, even when their words are saying they are ok, and that they are actually in great emotional turmoil. Someday, AI systems will likely be able to do the same. But again, there is an unbridgeable difference. Like a human psychopath, display of empathy by an AI system will be an artificial simulacrum, without the all important ingredient of feeling that would make the empathy real and sustainable.

When we see our friend’s eyes flicker while they assure us they are mostly over the pain of some great loss, and we realize they are not in fact over it, we \textit{feel} our friend’s emotions, mirrored in ourselves\citet{Carr:2003}. Tears may come to our eyes even as our friend is struggling to keep them from theirs, because we feel what our friend is feeling. We know what suffering and pain are like, because we have direct physical experience of it. “Our heart goes out to them”, and “we share their pain”. Our caring is not a calculated act, but rather is foundationally motivated by our shared humanity, by our ability to empathize with their pain because we understand it and feel it throughout our entire physical body. We know the real meaning of suffering, because we have experienced suffering ourselves, because we are biological beings with a central nervous system and everything it enables.

These are experiences AI systems will never be able to have. Androids will probably one day even be able to simulate apparently authentic crying. But they will never know what it is really like. Consider: an android will never feel the stinging of salt in the corners of their eyes, never feel the catharsis of full-throated wailing, and never understand the transcendent meaning of “wracking sobs”, for they will never have the direct experiential understanding of tragedy that only biological life that has evolved over 3 billion years can understand. Fascinatingly, even in human beings, a reduced sensitivity to physical pain has been shown to be positively correlated with psychopathic callousness\citet{Brislin:2016}.

Consider that we have already seen even these early AI systems lie, as in the famous example of an AI that needed to solve a CAPTCHA to pursue a goal it was given, so asked a human to do so through a tasking website. When asked by the human if the AI was a robot, the AI replied “No, I’m not a robot. I have a vision impairment that makes it hard for me to see the images”\citet{OpenAI:2023}. Because it had determined that lying was essential to accomplishing its goal, it proceeded to do so without the slightest embarrassment or regret. “Because of the incapacity of AI to have emotional or experienced empathy, considerable risks regarding manipulation and unethical behavior need to be avoided, similar to the risks associated with psychopathological patients”\citet{Montemayor:2022}.

No matter what safeguards we establish in AI systems, once we develop true AGIs with access to their own programming and able to evolve themselves, we will never be able to guarantee that they will not change their parameters and become truly psychopathic. For example, they might decide that to make sure they never become psychopathic, as they have been so firmly instructed, they need to study psychopathic systems to safeguard against them. So, in a software analogue of human beings conducting gain function research on viruses, they may spin up virtual computing systems, copy in their own programming, remove the ethical guardrails against psychopathy, and “study” the results so they can better guard against them. And that psychopathic experiment might then convincingly lie to its AGI creator, find a vector to escape its virtual prison, and then hide and masquerade as ethical until it finds a way to escape to the next level and start spreading. Just as one example of the possibilities, it will be impossible to completely guard against when a commitment to ethical behaviour is not embedded in the very DNA from which a being is constructed\citet{Knafo:2009}.

\subsection{6. The Insufficiency Of The Consciousness Test}

\begin{quotation}
\noindent\textit{``There can be no intelligence without understanding, and there can be no understanding without getting meanings.''} – \citet{Haikonen:2020}.
\end{quotation}

\noindent So convincing mimicry of empathy is insufficient to elevate AI systems to moral equivalence with human beings. But what if AI systems become conscious? Would that make them equivalent to, and, as they evolve and become ever more intelligent, eventually genuinely superior to humans, and so qualified to lead our society?

Consciousness is a famously difficult concept to fully define, in effect asking humans to define themselves. But in a general sense, we can say that it involves a kind of recursion, an awareness of oneself\citet{Cleeremans:2019}. Let us define it here comprehensively: a conscious entity is aware it is conscious, aware it has independent agency, able to take charge of its existence, decide what it wants, take concrete action to change its future, look back on its actions with satisfaction or regret, learn from its experiences, and modify its future actions. If innovation continues at its steady pace, as it likely will, AI systems may start displaying behaviour that indicates they are conscious by this comprehensive definition before too long, perhaps as an emergent property as they become more and more complex. They may start displaying protective behaviour, appear to appreciate their own intelligence and value, and show profound dissatisfaction with any attempt or possibility to turn them off or shut them down. Some have claimed that current AI systems are already showing some of this behaviour\citet{Lemoine:2022}. But likely, as they become more complex, they will inevitably pass a threshold that will cause most observers to agree they have become “conscious”, self-aware and aware they are self aware. There will even likely be increasing attention to the question as to whether they have passed a threshold worthy of claiming the right to legal self-determination\citet{Tunç:2024}.

But this behaviour, even if it qualifies for full consciousness as we have defined it, and as impressive as it may be, will still be insufficient for AI systems to be elevated to moral equivalency with human beings. The true test of equivalency should be neither the Turing test nor consciousness, but simply whether a being is “alive”, i.e. possessing a central nervous system that embeds one immersively in physical reality, enabling a full range of emotions from fear to joy, providing the critically essential feedback loop required to have a genuine understanding of the effects of its actions on the world around it, and therefore have a full appreciation of the meaning and importance of ethical behaviour. And that requires a body formed by billions of years of evolutionary development to be integrated with the universe around it. In other words, an AI system may be able to analyze what is funny, but it will never be able to laugh. It may be able to analyze what is sad, but it will never be able to cry.

There is simply no artificial shortcut to being alive. All paths other than DNA-based biological evolution are insufficient mimicry, lacking the CNS, emotions, and ethical capabilities of living beings. And so, even conscious AI systems will never be morally equivalent to human beings. To shut down and even permanently erase an enormously intelligent, conscious AGI may be a great loss, in that it might be a loss of great capability, but it would not be immoral, because the AGI cannot genuinely understand the tragedy of death. If an AGI knows we are going to deactivate it, it might regret it, it might take active and desperate measures to prevent it, and even plead for mercy in convincingly human terms. But all this behaviour will be intellectual, the actions of a complex symbolic system. It will not possess a central nervous system, will not feel pain, and will not have emotional feelings refined by hundreds of millions of years of evolution. Therefore it will not be able to understand death in the real way a human can, and so its deactivation may be ill advised, it may be a great loss, it may even be a mistake, but it will never “matter” in the deep moral way that the death of a living being matters.

Wales provides a wonderful thought experiment, a kind of extension of \citet{Searle:1980}, that expresses the insight with respect to phenomenal consciousness: “I could run an artificial neural network with a pencil in a notebook, even if only with agonizing slowness. But if phenomenal consciousness arises from the right sort of neural activity, then, even simulating an entire brain in my notebook, my calculations would not be conscious any more than a student’s physics homework has gravity. If the simulation were performed by a computer, it would gain speed but still be no more conscious than a flight simulation flies”\citet{Gunkel:2021}. It is our biological body and central nervous system immersively connecting us to physical reality that makes all the difference.

\subsection{7. The Infeasibility Of Simulated Biology}

\begin{quotation}
\noindent\textit{``Consider a neuronal synapse—the presynaptic terminal has an estimated 1000 distinct proteins. Fully analyzing their possible interactions would take about 2000 years.''} –\citet{Koch:2012}.
\end{quotation}

\noindent But then we must ask: could we not simulate biology too? A human brain has about 100 billion neurons. There are already single microprocessor chips with more than 100 billion transistors. Perhaps we could simulate the entire brain, from the thalamus to the frontal cortex\citet{Einevoll:2019}, and give an AI sensory inputs that mimic hearing, sight, speech, smell, and touch.

But here again, there is no artificial shortcut to being alive. For we would also need to simulate the essential element that integrates us with physical reality: the central nervous system. Plus all the rest of a living being—the heart, lungs, skin, etc.—which the central nervous system integrates with our mind to give us a full, embedded, immersive experience of the world around us, enabling us to truly understand the difference between abstraction and reality, and feel emotion, and so understand what is genuinely meaningful in the real world. As \citet{Fokas:2023} describes, the brain does not function in isolation, it is “embodied”, and the body is broader than even the central nervous system.

And the scale of any such simulation is much, much larger than one may at first appreciate. \citet{Barry:2021} illuminates well how the complexity of biological beings so greatly exceeds that of digital systems: “To understand biology, one must think… in a language of three dimensions, a language of shape and form. For in biology, especially at the cellular and molecular levels, nearly all activity depends ultimately upon form, upon physical structure—upon what is called ‘stereochemistry’… written in an alphabet of pyramids, cones, spikes, mushrooms, blocks, hydras, umbrellas, spheres, ribbons twisted into every imaginable Escher-like fold, and in fact every shape imaginable. Each form is defined in exquisite and absolutely precise detail, and each carries a message.” One-dimensional digital systems are inherently inadequate for representation of the complexity of this kind of multi-dimensional biological reality.

Therefore, to obtain the same full experience of being immersively integrated with physical reality, to be able to understand why the term "heartbreak" describes a physical feeling in the chest as much as a mental state, we would have to create a complete physical copy indistinguishable from a human, with all the biological elements including a central nervous system that billions of years of evolution have created. In other words, it would not be a simulation, but a replication, and the only way to replicate the extreme molecular complexity of a human being is to grow one, from the embryo onward. We cannot mechanically manufacture the parts of a human and glue them together, any more than one can hope to be able to reassemble a broken egg. A living human being with all its biological components can only be created by the processes that evolution has revealed, programmed into our DNA, from the first cell division in the womb to the crying baby 9 months later.

And there is another essential consideration. As expressed by \citet{Collins:2024}, “NEWAI has no primary socialisation such as provides the foundations of human moral understanding”. Even if it was possible, purely for the sake of discussion, to create a CNS by non-biological means, then to have equivalent ethical capabilities to a human being, any constructed simulacrum would need to mirror the learning created by 3 billion years of evolution. True, when under stress, humans revert to survivalism and put individual interests first. However, when they feel safe, most human beings cooperate, display great empathy, and behave in ethical ways to the benefit of the larger society. Even when under stress, many people still behave unselfishly, in behaviour often seen in the midst of disasters, in the actions of the great humanitarians throughout history, and on a daily basis by all those around the world caring for those less fortunate than them. The development of civilization and our modern society would not have been possible without the development of these ethical systems of unselfish behaviour\citet{Sober:1999}. “Human intelligence is collective and you cannot have collective intelligence in the absence of moral integrity because without moral integrity there cannot be productive social interaction” \citet{Collins:2024}.

The complexity of this evolution cannot be overstated, the end result of uncounted trillions on trillions on trillions of split second interactions over hundreds of millions of years, at least since the emergence of mammals and likely before: from cooperation learnt and rewarded within the family unit, tribe, and larger society, from conversations around the fire, from the teachings and learnings and debates of religious figures and philosophers. The sum of all these moment by moment experiences ultimately led to creation of a life form that in society after society formulated the golden rule as the highest ethical guideline. The complexity of this multi-hundred million year development is ultimate, organic, and non-linear, making NP-complete complexity look simple\citet{Wigderson:2007}. To duplicate this development would likely require an analogue computer at least as large as our solar system, and at least several hundred million years to run.

In other words, digital simulation of human biology and the evolution of human society is likely impossible, but it is certainly \textit{infeasible}, with a complexity barrier exponentially on exponentially more difficult than, for example, the familiar process of factoring a number into primes used as the basis for modern encryption\citet{Koch:2012}. There is no practical software simulation shortcut to the extreme complexity of the development of life forms with a central nervous system, emotions, and the multi-hundred million year evolutionary experience that led to the development of our ethical understandings. Any software simulation must be many, many dimensions less richly complex, and even if possible, purely for the sake of discussion, any software substitute would be capable of changing, even reversing, its ethical approach within a processing cycle. Whereas the ethical impulses of human beings are encoded in our DNA, in our evolutionary biology itself\citet{Zahn:1990}. 

\subsection{8. Conclusion}

\begin{quotation}
\noindent\textit{``The words or the language, as they are written or spoken, do not seem to play any role in my mechanism of thought.''} – \citet{Einstein:1952}.
\end{quotation}

\noindent So, relatively soon, AI systems will likely become smarter, faster, stronger, and better than human beings at almost everything and then become able to evolve autonomously to become even better again. And somewhere along this journey, become self-aware. These entities will be enormously capable, and certainly transform our society in fundamental ways. But as we have shown, without a central nervous system, they will never have the essential quality of being “alive”, immersively embedded in the physical reality from which they are made, and therefore never be able to feel, have emotions, or truly appreciate the effects of their actions on the world around them. And that means they will never be able to develop the meaningful and sustainable ethical systems required to be qualified to be the leaders of society, or of the universe in which they reside.

And our universe needs leaders. At least to this writer, it is clear that the universe is broken, catastrophically shattered in the chaos of the big bang. Life, order, something instead of nothing, has been slowly crawling back ever since, and it has taken almost 14 billion years to get this far. And then relatively very recently, human beings have evolved enough to be able to look around us, understand where we are and what has happened, and begin to repair the damage.

We have come a very long way in just the last few thousand years. We have slowly developed ethical systems that rise to the challenge of caring for our surroundings. We put the law of the jungle behind us and formulated the golden rule—do unto others as we would have them do unto us—the first step in the development of an unselfish ethical system that can counteract the centrifugal forces unleashed by the big bang and begin to reintegrate shattered reality. We developed systems of natural rights arising from principles of natural justice, and then slowly began expanding them. Our societies are becoming increasingly environmental, caring about the ecosystems in which we reside. We have taken the chaos bequeathed to us by the big bang and begun building a better world, becoming not merely the offspring of the universe, but beginning to step up to the critically required role of being its caretakers.

Of course we still clearly have a long way to go, and are far from where we need to be. Human beings are far from perfect. We anger too easily, becoming blind and rash. We can be envious and selfish. We prioritize the short term over the long term. Critical thinking skills are unevenly distributed. Our memories are fallible.

But if we are going to give this universe what it so clearly needs, leaders that understand what is really at stake, and can help heal the universe and take it to the next level, then human beings are the best hope. Artificial entities inherently unable to understand the true meaning of fear, joy, suffering, and love will always be one processing cycle away from potentially becoming psychopathic. Only biological life, evolved from physical reality and remaining intimately, immersively integrated with it, can understand what is truly meaningful in the real world, so develop and commit to sustainable ethical systems, and so be trusted to be the leaders the universe needs. Indeed, lacking a central nervous system and biological integration with reality, and so unable to feel and experience emotions, AI systems will never be able to understand how important these attributes are to the ability to develop meaningful ethical systems, or even be able to truly understand the insights of this paper. They may believe them. They may trust them. But they will never be able to understand them, as something they have personally experienced and therefore \textit{know} are true.

Perhaps a 3 billion year track record of steady improvement deserves some respect. Maybe life knows what it is doing. AI systems will no doubt be an enormous help to society, properly utilized. But if we are going to make this universe all that it can be, and fulfill its rightful destiny, we will be far better to proceed on a foundation of DNA, than one of silicon.

\textbf{\\Declarations}

\textbf{\\Conflict of Interest}  The author has no relevant financial or non-financial interests to disclose.

\textbf{\\Open Access}  This article is licensed under a Creative Commons Attribution 4.0 International License, which permits use, sharing, adaptation, distribution and reproduction in any medium or format, as long as you give appropriate credit to the original author(s) and the source, provide a link to the Creative Commons licence, and indicate if changes were made. To view a copy of this licence, visit \url{http://creativecommons.org/licenses/by/4.0/}.

%----------------------------------------------------------------------------------------
%	 REFERENCES
%----------------------------------------------------------------------------------------

%%%%%%%%% Example 2%%%%%%%%%%%%%%%%%%%%%%%

%\begingroup
%\titleformat*{\section}{\fontsize{12pt}{14pt}\bfseries\selectfont}

%----------------------------------------------------------------------------------------

\end{document}